\documentclass{article}

\PassOptionsToPackage{numbers, compress,sort}{natbib}

\usepackage[final]{neurips_2020}

\usepackage[utf8]{inputenc} 
\usepackage[T1]{fontenc}    
\usepackage{url}            
\usepackage{booktabs}       
\usepackage{amsfonts}       
\usepackage{nicefrac}       
\usepackage{microtype}      
\usepackage[pagebackref=true,breaklinks=true,letterpaper=true,colorlinks,bookmarks=false]{hyperref}

\usepackage{graphicx}
\usepackage{subfig}
\usepackage{multirow}
\usepackage{bbm}
\usepackage{color}
\usepackage{arydshln}

\usepackage{makecell}
\usepackage{authblk}
\title{Self-Supervised Ranking for Representation Learning}

\author[1]{Ali Varamesh}
\author[1]{Ali Diba}
\author[1]{Tinne Tuytelaars}
\author[1,2]{Luc Van Gool}

\affil[1]{\footnotesize ESAT-PSI, KU Leuven} 
\affil[2]{\footnotesize CVL, ETH Zurich}
\affil[ ]{\textit {\{firstname.lastname\}@esat.kuleuven.be }}
\begin{document}

\maketitle

\begin{abstract}

We present a new framework for self-supervised representation learning by formulating it as a ranking problem in an image retrieval context on a large number of random views (augmentations) obtained from images. Our work is based on two intuitions: first, a good representation of images must yield a high-quality image ranking in a retrieval task; second, we would expect random views of an image to be ranked closer to a reference view of that image than random views of other images. Hence, we model representation learning as a learning to rank problem for image retrieval. We train a representation encoder by maximizing average precision (AP) for ranking, where random views of an image are considered positively related, and that of the other images considered negatives. The new framework, dubbed S2R2, enables computing a global objective on multiple views, compared to the local objective in the popular contrastive learning framework, which is calculated on pairs of views. In principle, by using a ranking criterion, we eliminate reliance on object-centric curated datasets. When trained on STL10 and MS-COCO, S2R2 outperforms SimCLR and the clustering-based contrastive learning model, SwAV, while being much simpler both conceptually and at implementation. On MS-COCO, S2R2 outperforms both SwAV and SimCLR with a larger margin than on STl10. This indicates that S2R2 is more effective on diverse scenes and could eliminate the need for an object-centric large training dataset for self-supervised representation learning.

 
\end{abstract}

\section{Introduction}

Self-supervised visual representation learning (SSRL) has advanced quickly, thanks to the contrastive learning principle and extensive research on the problem. Contrastive learning has a simple, intuitive objective: to assure that similar images are mapped to a compact neighborhood in the representation space. In practice, contrastive representation learning methods operate on pairs of views extracted from images, and implicitly, they assume that there is a dominant object that defines each image. Based on this assumption, then either image instance ID \cite{chen2020simple,he2020momentum,tian2019contrastive,wu2018unsupervised,misra2020self,grill2020bootstrap} or a simultaneously generated cluster code for each image \cite{caron2020unsupervised,asano2019self,caron2018deep,li2020prototypical} is used to define a cross-entropy objective for training an encoder such that pairs of augmentations from an image are classified as the same compared to a set of negative views obtained from other images. However, in a real-world scene, it is natural to have very different objects at different spatial locations \cite{purushwalkam2020demystifying}, and forcing every possible pair of views to have the same representation in isolation would not be reasonable. 

We propose a new framework for SSRL that naturally lends itself to compare and contrast a large number of views from images. In the new framework, S2R2, we formulate representation learning as a retrieval task with the objective of optimizing the ranking of random views of images in terms of average precision. The ranking is a stronger criterion than the contrastive learning loss, as it goes beyond a local objective by operating on multiple views. With a local objective that simply forces pairs of positive samples to be close to each other and far from the negatives, it is inevitable to regularly pull together a pair of positive samples while moving either one away from other related samples or by pulling them closer to the space of negative samples. This issue potentially slows down the optimization process. In contrast, by simultaneously ranking a large number of views, the global objective adapted in S2R2 avoids such conflicting optimization steps. The issue of locality in contrastive learning is illustrated in \cite{varamesh2020mix} by showing that representations learned by SimCLR \cite{chen2020simple} are not semantically disentangled enough, and a mixture model is deployed to enforce disentanglement at the category level. 


A remedy for the locality issue could be extending the pairwise contrastive objective to use more crops for computing the loss, a technique adopted by a recent state-of-the-art model, SwAV \cite{caron2020unsupervised}. This technique makes SwAV more efficient than its more basic counterpart, SimCLR \cite{chen2020simple}. Still, it is not able to match the ranking objective in terms of performance. Technically the multi-crop technique is not a global objective; rather, it accumulates loss for multiple pairs. Moreover, SwAV relies on a multi-stage algorithm that needs solving a cluster assignment problem at every optimization step. 

We summarise our contributions as follows:
\begin{itemize}
    \item We propose a new framework, S2R2, for self-supervised representation learning by ranking random image views. S2R2 employs a global optimization objective and fundamentally does not rely on object-centric curated images.
    \item We empirically show that S2R2 outperforms the state-of-the-art contrastive learning models SimCLR and SwAV.
\end{itemize}


\section{Methodology}

We first review the definition of ranking in image retrieval and how to maximize ranking average precision (AP) end-to-end. Then we proceed to describe S2R2 in terms of maximizing ranking AP.

In ranking based image retrieval, given a query image $I_q$ and a set of $m$ images $I = \{I_1 ... I_m\}$ to search, an ideal ranking algorithm should rank the set of all relevant (positive) images $I_P$ above the set of unrelated (negative) images $I_N$, where  $I= I_P \cup I_N$ and $I_P \cap I_N = \emptyset$. For image retrieval, AP is the standard metric to measure the quality of a ranking algorithm. Following \cite{qin2010general} and \cite{brown2020smooth}, the definition of AP is shown in eq. \ref{equ:ap_def}, where the function $R_q(i,X)$ ranks image $I_i$ among all images in $X$ with respect to the query image $I_q$. AP is maximized once all positive images are ranked higher than any negative image.

\begin{equation}
       AP_q = \frac{1}{|I_P|} \sum_{i \in I_P} \frac{R_q(i, I_P) }{ R_q(i,I)}
\label{equ:ap_def}
\end{equation}

The rank function $R_q(i, X)$ returns the number of images in $X$ that are more similar to the query image $I_q$ than image $I_i$. This is formally defined in eq. \ref{equ:rank_fn_def}, where $s_{u,v}$ denotes the cosine similarity between images $I_u$ and $I_v$ based on their representation vectors $r_u$ and $r_v$, respectively. Although we use cosine similarity in this work, one may use any standard similarity metric for computing $s_{u,v}$. 

\begin{equation}
       R_q(i,X) = 1+ \sum_{j \in X, j\neq i} \mathbbm{1}{\{(s_{q,j} - s_{q,i})>0\}} \qquad s.t. \qquad s_{u,v} = \langle  \frac{r_u}{||r_u||}.\frac{r_v}{||r_v||} \rangle
\label{equ:rank_fn_def}
\end{equation}


Unlike the contrastive learning objective, average precision does not force all positive images to have the same representation; neither it expects them all to have the same identity. Instead, the goal is, given a query image, to assure all the positive images are closer to the query than any negative image.

Optimizing the AP loss above is not trivial, as it is not differentiable due to the presence of the indicator function $\mathbbm{1}{\{\}}$. However, recently Brown et al.~\cite{brown2020smooth} have proposed a simple modification to this formulation that results in a highly accurate approximation of eq.~\ref{equ:rank_fn_def}. That is, by replacing the indicator function with the logistic sigmoid function shown in eq.~\ref{equ:sigmoid}, where $\tau$ is a temperature parameter. With this modification, eq.~\ref{equ:ap_def} becomes differentiable, and we can train a ranking model by directly maximizing AP. Brown et al. \cite{brown2020smooth} report significant and robust improvements for image retrieval in various settings using this approximation. 
\begin{equation}
       \phi (d; \tau) = \frac{1}{1 + e^{-d/\tau}}
\label{equ:sigmoid}
\end{equation}

In S2R2, we adapt this objective and maximize AP by simulating a retrieval setting where the positive samples are obtained by applying aggressive augmentation to an image. Similarly, negatives are obtained by applying similar augmentations to other images. 
In our mini-batch gradient descent optimization setting, first, we sample $B$ images and then generate $K$ views for each image by drawing an augmentation sequence for each view. We use the same aggressive augmentation setup as in SimCLR\cite{chen2020simple}. At every SGD step, each of the $B*K$ views is once used as a query image; with the other $k-1$ views from the same reference image constituting positive samples ($|I_P|= k-1$), and the $k*(B-1)$ views of the other images forming negatives ($|I_N|= k*(B-1)$). The final optimization target is obtained by averaging the AP in eq.~\ref{equ:ap_def} computed for rankings w.r.t to each of the $B*K$ query images.

{\setlength{\extrarowheight}{2pt}%
\begin{table}[h]
\caption{Top-1 accuracy on STL10 for representation encoders trained with ResNet18}
\begin{center}
\resizebox{.8\textwidth}{!}{
\begin{tabular}{c|c|cc|cccccccc}
 & &\multicolumn{2}{c}{\thead{\textbf{Setting}}} &  \multicolumn{8}{c}{\textbf{Epochs}} \\

 \multicolumn{1}{c|}{\thead{\textbf{Training data}}} &\multicolumn{1}{c}{\thead{\textbf{Model}}} & \multicolumn{1}{|c}{\thead{\textbf{\#images}}} & \multicolumn{1}{c|}{\thead{\textbf{\#views}}} & \textbf{10} & \textbf{50} & \textbf{100} & \textbf{200} & \textbf{400} & \textbf{600} & \textbf{800} & \textbf{1000} \\ \Xhline{4\arrayrulewidth} 

\multirow{3}{*}{\thead{STL10 \\(unlabeled + train)}} & SimCLR & 256 & 2& $64.2$ & $74.3$ & $77.8$ & $82.0$  & $84.4$ & $86.1$ & $84.9$ & $86.1$\\

& SwAV & 256 & 8 &  $67.4$ & $79.3$ & $82.8$ & $85.6$ & $87.4$ &$88.4$ & $89.5$ & $89.4$\\


& S2R2 & 64 & 20 & $72.4$ & $81.6$& $84.1$ & $86.4$& $88.4$ & $89.1$ & $89.4$ & $89.8$\\

\Xhline{2\arrayrulewidth} 
\multirow{3}{*}{COCO-Train} & SimCLR & 256 & 2 & $60.7$ & $68.6$ & $70.9$ & $73.4$ & $75.1$ &  $74.9$ & $74.1$ & $74.5$\\

&  SwAV & 128 & 6 & $65.2$ & $71.7$ & $75.0$ & $75.4$ & $77.1$ & $77.0$& 77.0 & 76.8 \\

&  S2R2 &  64 & 20 & $68.3$ & $74.2$ & $76.2$ & $77.2$ & $77.4$ & $77.5$ & $78.0$ & $78.5$ \\
\Xhline{2\arrayrulewidth} 

\multirow{3}{*}{COCO-Val}  &  SimCLR  &64 & 2 & $46.1$ & $56.1$ & $58.0$ & $62.0$ & $63.5$ & $64.1$ & $64.4$ & $64.8$\\

 &  SwAV  & 32 & 40 & $44.1$ & $59.7$ & $63.7$ & $66.3$ & $66.3$ & $65.0$ & $63.3$ & $62.8$ \\

&  S2R2 & 8 & 10 &  $58.6$ & $64.8$ & $67.5$ & $69.0$ & $ 68.8$ & $68.8$ & $67.3$ & $68.5$  \\
\end{tabular}
}
\end{center}
\label{fig:_evals_res18}
\end{table}
}

\section{Experiments}
We compare S2R2 to two recent state-of-the-art models, SimCLR \cite{chen2020simple} and SwAV \cite{caron2020unsupervised}, on object-centric and cluttered datasets. For a cluttered dataset, we use the MS-COCO dataset \cite{lin2014microsoft}. To analyze all models' behavior in response to dataset size on MS-COCO, we separately train models on its train split (COCO-Train) that includes about 118k images, and validation split (COCO-Val) that has only 5K images. For an object-centric curated dataset, we use STL10 \cite{coates2011analysis}, a subset of ImageNet designed specifically for research on unsupervised methods. It includes 100k unlabeled images from 10 known and some unknown categories. It also has labeled train and test splits (STL10-Train and STL10-Test), with 5k and 8k images, respectively, from the ten known categories. For STL10, we use the combination of the unlabeled and train splits (105k images) for training representation encoders.

To evaluate a given representation encoder (trained on either MS-COCO or STL10), we adopt the standard linear classification setting \cite{wu2018unsupervised}. A linear classifier is trained on STL10-Train using the frozen (detached) representations by an encoder, and the Top-1 accuracy on STL10-Test is reported.



Given that we have access to limited computational infrastructure, we resize the random image views to 96x96 (the original size for STL10 images) for training on all datasets. We experiment with both ResNet18 and ResNet50 convolutional backbones \cite{he2016deep}. Following the best practice in SSRL \cite{chen2020simple}, at training, we use a non-linear projection layer on top of the representations. All models in this paper are trained from scratch, and we report the best performance we achieve for each model (including SimCLR and SwAV) by searching for the best hyperparameters. For SwAV, especially, we have tried to find the best multi-crop setting for each dataset separately. For S2R2, we use ADAM with a fixed learning rate of $1e-4$; for SimCLR, we use ADAM with an initial learning rate of $3e-4$ and cosine decay, and for SwAV, we use the LARS optimizer with an initial learning rate of $1$ and cosine decay.


Tables \ref{fig:_evals_res18} and \ref{fig:_evals_res50} show linear classification accuracy for each model when trained on STL10, COCO-Train, and COCO-Val. On COCO-Train and COCO-Val, S2R2 outperforms both SimCLR and SwAV by a large margin. The trend is similar on STL10, however, with a smaller margin. This indicates that S2R2 is more effective on cluttered datasets. We believe computing a global objective on multiple views makes ranking a better objective on un-curated datasets like MC-COCO. Interestingly, on the small COCO-Val split, S2R2 converges at a much faster rate. We also observe that, surprisingly, the performance of SwAV degrades when trained with a longer schedule, which demands further investigation. Note that the close performance of S2R2 and SwAV on STL10 is not surprising, as the contrastive loss already suits the best that setting. Besides, SwAV deploys a multi-crop technique, which is enough to mimic a global metric on a curated dataset. 


{\setlength{\extrarowheight}{2pt}%
\begin{table}[h]
\caption{Top-1 accuracy on STL10 for representation encoders trained with ResNet50}
\begin{center}
\resizebox{.6\textwidth}{!}{
\begin{tabular}{c|c|cc|cccc}
 & &\multicolumn{2}{c}{\thead{\textbf{Setting}}} &  \multicolumn{4}{c}{\textbf{Epochs}} \\

 \multicolumn{1}{c|}{\thead{\textbf{Training data}}} &\multicolumn{1}{c}{\thead{\textbf{Model}}} & \multicolumn{1}{|c}{\thead{\textbf{\#images}}} & \multicolumn{1}{c|}{\thead{\textbf{\#views}}} & \textbf{10} & \textbf{50} & \textbf{100} & \textbf{200} \\ \Xhline{4\arrayrulewidth} 

\multirow{2}{*}{\thead{STL10 \\(unlabeled + train)}} & SimCLR & 256 & 2& $63.7$ & $76.6$ & $81.1$ & $85.2$\\
 & S2R2 & 32 & 20 & $77.3$ & $85.9$ & $88.4$ & $89.7$\\
 \Xhline{2\arrayrulewidth} 
 
\multirow{2}{*}{COCO-Train} & SimCLR & 256 & 2 & $60.4$ & $68.1$ & $74.9$ & $77.5$\\
 & S2R2 & 32 & 20 & $74.1$ & $79.7$ & $81.2$ & $80.1$\\
\Xhline{2\arrayrulewidth} 
 
 \multirow{2}{*}{COCO-Val} & SimCLR & 256 & 2 & $39.0$ & $48.0$ & $53.4$ & $59.4$\\
 & S2R2 & 16 & 20 & $54.3$ & $65.6$ & $68.4$ & $70.2$\\
\end{tabular}
}
\end{center}
\label{fig:_evals_res50}
\end{table}
}

\begin{figure}[t]
\centering

\subfloat[COCO-Train]{\label{fig:speedaccfull}\includegraphics[width=.33\textwidth]{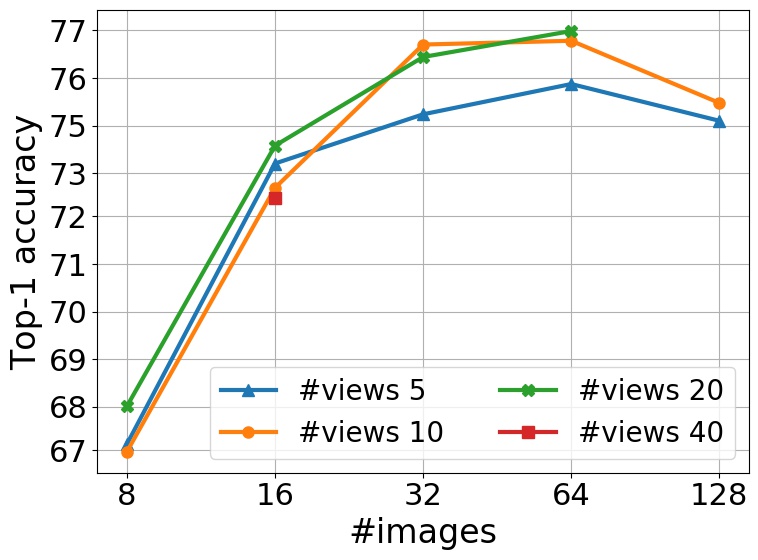}}
\subfloat[COCO-Val]{\label{fig:speedacc10}\includegraphics[width=.33\textwidth]{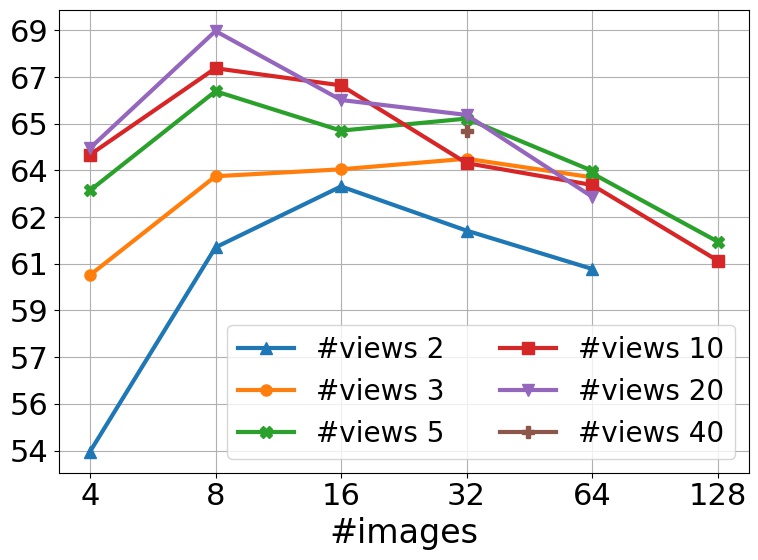}}
\subfloat[STL10]{\label{fig:speedacc10}\includegraphics[width=.33\textwidth]{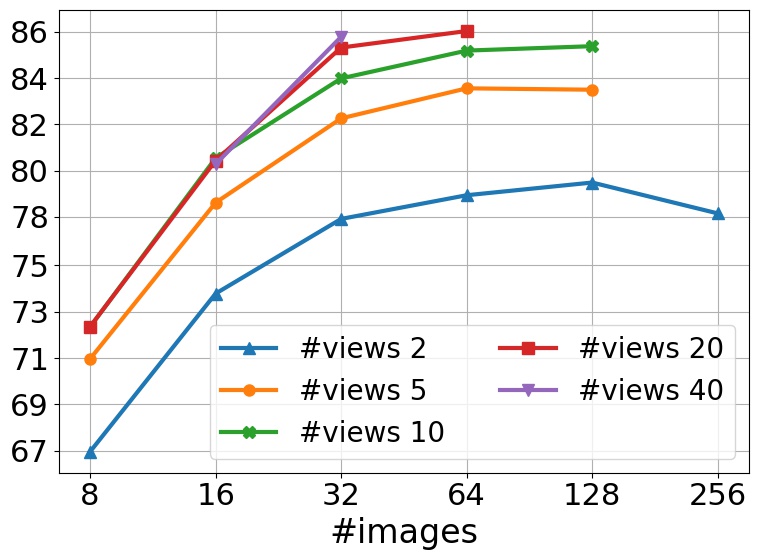}}

\caption{Linear classification accuracy v.s. the number of images/views in a mini-batch (ResNet18)}
\label{fig:batchsize_ablaton}
\end{figure}

Figure \ref{fig:batchsize_ablaton} shows the effect of varying the number of images and the views per image in a mini-batch on the performance of S2R2 for different datasets. As we can see, increasing the number of images not only does not help beyond a point but also degrades accuracy. We believe this happens at the point where images in a mini-batch are likely to become similar with respect to their visual content. Therefore, their random views can not be ranked properly. Nevertheless, this demands more investigation by controlling the dataset content in terms of visual diversity (e.g. by controlling the number of categories). On the other hand, the mini-batch size should be larger than a threshold, which depends on the dataset size and complexity, as the figure indicates.

\section{Discussion}

We show that our new ranking based framework for self-supervised representation learning, S2R2, outperforms contrastive learning methods of SimCLR and SwAV. S2R2 shows impressive results, particularly on cluttered datasets. In this work, we experimented with MS-COCO as a cluttered dataset; however, it is not an ideal dataset to represent crowded real-world scenes as it is heavily biased towards the person class; more than half the images include humans. Moreover, most images are selected to cover objects at different scales, not necessarily bringing in more diverse visual content. For example, it comes with a large number of baseball scenes that are extremely similar in terms of visual complexity. We plan to run experiments on more challenging datasets like Places \cite{zhou2017places} or ADE20k \cite{zhou2017scene} to better understand the performance gains of S2R2 compared to SimCLR or SwAV. Similarly, in the future, we would like to train S2R2 on the ImageNet \cite{deng2009imagenet} in order to have a more comprehensive comparison to the state-of-the-art in self-supervised representation learning.

For the augmentation procedure, in this work, we only experimented with uniformly resized and scaled image crops; however, a more sophisticated cropping policy could be used to ensure covering all spatial locations at different scales, which needs further experiments. Finally, S2R2 outperforms contrastive methods; however, it does not solve the problem of potentially very similar views of different images being considered as negatives, an issue that demands further research.


\bibliography{neurips_2020}

\begin{thebibliography}{10}\itemsep=-1pt

\bibitem{brown2020smooth}
Andrew Brown, Weidi Xie, Vicky Kalogeiton, and Andrew Zisserman.
\newblock Smooth-ap: Smoothing the path towards large-scale image retrieval.
\newblock {\em arXiv preprint arXiv:2007.12163}, 2020.

\bibitem{caron2018deep}
Mathilde Caron, Piotr Bojanowski, Armand Joulin, and Matthijs Douze.
\newblock Deep clustering for unsupervised learning of visual features.
\newblock In {\em Proceedings of the European Conference on Computer Vision
  (ECCV)}, pages 132--149, 2018.

\bibitem{caron2020unsupervised}
Mathilde Caron, Ishan Misra, Julien Mairal, Priya Goyal, Piotr Bojanowski, and
  Armand Joulin.
\newblock Unsupervised learning of visual features by contrasting cluster
  assignments.
\newblock {\em arXiv preprint arXiv:2006.09882}, 2020.

\bibitem{chen2020simple}
Ting Chen, Simon Kornblith, Mohammad Norouzi, and Geoffrey Hinton.
\newblock A simple framework for contrastive learning of visual
  representations.
\newblock {\em arXiv preprint arXiv:2002.05709}, 2020.

\bibitem{coates2011analysis}
Adam Coates, Andrew Ng, and Honglak Lee.
\newblock An analysis of single-layer networks in unsupervised feature
  learning.
\newblock In {\em Proceedings of the fourteenth international conference on
  artificial intelligence and statistics}, pages 215--223, 2011.

\bibitem{deng2009imagenet}
Jia Deng, Wei Dong, Richard Socher, Li-Jia Li, Kai Li, and Li Fei-Fei.
\newblock Imagenet: A large-scale hierarchical image database.
\newblock In {\em 2009 IEEE conference on computer vision and pattern
  recognition}, pages 248--255. Ieee, 2009.

\bibitem{grill2020bootstrap}
Jean-Bastien Grill, Florian Strub, Florent Altch{\'e}, Corentin Tallec,
  Pierre~H Richemond, Elena Buchatskaya, Carl Doersch, Bernardo~Avila Pires,
  Zhaohan~Daniel Guo, Mohammad~Gheshlaghi Azar, et~al.
\newblock Bootstrap your own latent: A new approach to self-supervised
  learning.
\newblock {\em arXiv preprint arXiv:2006.07733}, 2020.

\bibitem{he2020momentum}
Kaiming He, Haoqi Fan, Yuxin Wu, Saining Xie, and Ross Girshick.
\newblock Momentum contrast for unsupervised visual representation learning.
\newblock In {\em Proceedings of the IEEE/CVF Conference on Computer Vision and
  Pattern Recognition}, pages 9729--9738, 2020.

\bibitem{he2016deep}
Kaiming He, Xiangyu Zhang, Shaoqing Ren, and Jian Sun.
\newblock Deep residual learning for image recognition.
\newblock In {\em Proceedings of the IEEE conference on computer vision and
  pattern recognition}, pages 770--778, 2016.

\bibitem{li2020prototypical}
Junnan Li, Pan Zhou, Caiming Xiong, Richard Socher, and Steven~CH Hoi.
\newblock Prototypical contrastive learning of unsupervised representations.
\newblock {\em arXiv preprint arXiv:2005.04966}, 2020.

\bibitem{lin2014microsoft}
Tsung-Yi Lin, Michael Maire, Serge Belongie, James Hays, Pietro Perona, Deva
  Ramanan, Piotr Doll{\'a}r, and C~Lawrence Zitnick.
\newblock Microsoft coco: Common objects in context.
\newblock In {\em European conference on computer vision}, pages 740--755.
  Springer, 2014.

\bibitem{misra2020self}
Ishan Misra and Laurens van~der Maaten.
\newblock Self-supervised learning of pretext-invariant representations.
\newblock In {\em Proceedings of the IEEE/CVF Conference on Computer Vision and
  Pattern Recognition}, pages 6707--6717, 2020.

\bibitem{purushwalkam2020demystifying}
Senthil Purushwalkam and Abhinav Gupta.
\newblock Demystifying contrastive self-supervised learning: Invariances,
  augmentations and dataset biases.
\newblock {\em arXiv preprint arXiv:2007.13916}, 2020.

\bibitem{qin2010general}
Tao Qin, Tie-Yan Liu, and Hang Li.
\newblock A general approximation framework for direct optimization of
  information retrieval measures.
\newblock {\em Information retrieval}, 13(4):375--397, 2010.

\bibitem{tian2019contrastive}
Yonglong Tian, Dilip Krishnan, and Phillip Isola.
\newblock Contrastive multiview coding.
\newblock {\em arXiv preprint arXiv:1906.05849}, 2019.

\bibitem{varamesh2020mix}
Ali Varamesh and Tinne Tuytelaars.
\newblock Mix'em: Unsupervised image classification using a mixture of
  embeddings.
\newblock {\em arXiv preprint arXiv:2007.09502}, 2020.

\bibitem{wu2018unsupervised}
Zhirong Wu, Yuanjun Xiong, Stella~X Yu, and Dahua Lin.
\newblock Unsupervised feature learning via non-parametric instance
  discrimination.
\newblock In {\em Proceedings of the IEEE Conference on Computer Vision and
  Pattern Recognition}, pages 3733--3742, 2018.

\bibitem{asano2019self}
Asano YM., Rupprecht C., and Vedaldi A.
\newblock Self-labelling via simultaneous clustering and representation
  learning.
\newblock In {\em International Conference on Learning Representations (ICLR)},
  2020.

\bibitem{zhou2017places}
Bolei Zhou, Agata Lapedriza, Aditya Khosla, Aude Oliva, and Antonio Torralba.
\newblock Places: A 10 million image database for scene recognition.
\newblock {\em IEEE Transactions on Pattern Analysis and Machine Intelligence},
  2017.

\bibitem{zhou2017scene}
Bolei Zhou, Hang Zhao, Xavier Puig, Sanja Fidler, Adela Barriuso, and Antonio
  Torralba.
\newblock Scene parsing through ade20k dataset.
\newblock In {\em Proceedings of the IEEE conference on computer vision and
  pattern recognition}, pages 633--641, 2017.

\end{thebibliography}
\bibliographystyle{neurips_2020}

\end{document}